\documentclass[11pt]{article}
\usepackage[margin=1in]{geometry}
\usepackage{amsmath,amssymb,amsthm,mathtools}
\usepackage{booktabs}
\usepackage{array}
\usepackage{enumitem}
\usepackage{microtype}
\usepackage[colorlinks=true,linkcolor=blue,citecolor=blue,urlcolor=blue]{hyperref}
\setlist[itemize]{leftmargin=1.2em,itemsep=2pt,topsep=3pt}
\setlist[enumerate]{leftmargin=1.4em,itemsep=2pt,topsep=3pt}
\newtheorem{theorem}{Theorem}
\newtheorem{lemma}{Lemma}
\newtheorem{proposition}{Proposition}

\theoremstyle{remark}
\newtheorem{remark}{Remark}
\newcommand{\E}{\mathbb{E}}
\newcommand{\R}{\mathbb{R}}
\newcommand{\tr}{\operatorname{tr}}
\newcommand{\rank}{\operatorname{rank}}
\newcommand{\Stab}{\operatorname{Stab}}
\newcommand{\Omat}{O(k)}
\title{\textbf{Same-State Pairs Make One JEPA Proof of the Platonic Representation Hypothesis Vacuous}\\[2pt]
\large A Correction and a Gauge Ledger}
\author{Xin Li\\
Shenzhen University of Advanced Technology\\
\small\texttt{suat25000420@stu.suat-sz.edu.cn}}
\date{November 2026}
\hypersetup{pdftitle={Same-State Pairs Make One JEPA Proof of the Platonic Representation Hypothesis Vacuous: A Correction and a Gauge Ledger},pdfauthor={Xin Li}}

\begin{document}
\maketitle

\begin{abstract}
In a Joint-Embedding Predictive Architecture (JEPA), the JEPA proof of the Platonic Representation Hypothesis (PRH) assumes that every observation determines the world state and that a spectral gap makes minimizers unique up to rotation. Under the hard-whitening population objective, same-state positive pairs make the induced world kernel the identity: every centered whitened state function is a global minimizer, and, when $d_z-1\ge 2k$, two minimizers can have best-rotation squared residual $2k$. The uniqueness claim is conditional; for $k<d_z-1$, where $d_z$ is the number of world states, this limit makes its premise unsatisfiable, so the claim has no content there. The experiments that motivate it leave the assumption: a caption sees only part of an image, and the paper states no experiment meets its hypotheses exactly. The theorem does not reach those experiments, and for $k<d_z-1$ its idealized same-state limit is vacuous. One of the paper's four quantitative results survives: a gap-free lower bound. Its bound vanishes only at perfect scene-explained fraction; it bounds alignment error but cannot certify good-view alignment. A canonical frame is determined only up to a rotation and a latent permutation. Paired samples fix the relative frame at effective rank $k$; a global rotation remains. Component semantics fix the relative permutation only up to signs; value anchors fix it once the mixing frame is pinned. A resource--gauge ledger records which resource fixes which gauge.
\end{abstract}

\section{Introduction}\label{sec:intro}

Consider two views of one scene in the paper's idealized model, where every view determines the world state. The induced positive-pair kernel is the identity: the objective can be driven to zero by any whitened function of the state, so two trained encoders can be orthogonal. This is not an optimization failure. It is what the idealized same-state model asks for.

The numbers are not marginal. With $|Z|=60$ world states and $k=8$ embedding dimensions, the minimizer set is a Stiefel manifold of dimension $436$, while a single rotation orbit has dimension $28$. When the state space is large enough, two global minimizers can have best-rotation squared residual $2k$, the maximum possible. For $k=3$, that is $6$ out of $6$.

This matters because the Platonic Representation Hypothesis (PRH) conjectures that representations converge as models scale \cite{huh2024}; a recent JEPA-based theorem states a precise version, agreement up to one rotation \cite{balestriero2026}.\footnote{This note comments on arXiv:2610.39302v1 (29 October 2026).}\footnote{An AI writing assistant was used in preparing this manuscript; the author is responsible for all content.} The theorem assumes injective renderings and that minimizers are unique up to rotation; its Proposition~1 gives this under a spectral gap. The experiments that motivate it leave the injective assumption: a caption sees only part of an image. The distinction between the idealized limit and the experiments matters for evaluation: agreement can be high because two encoders share a subspace while the rotation between them stays a free gauge, so an agreement number computed through a fitted rotation certifies an aligning rotation, not the uniqueness assumption on which the theorem rests. The question is whether the theorem applies to the evidence it is meant to explain, and what information would make ``identifiable up to a rotation'' a scientific statement. This note separates the two questions. It offers a correction and a resource--gauge ledger, not a new convergence theorem.

The paper's formal object is more general than its motivating sentence. Appendix~A.4 of \cite{balestriero2026} lets good views be any symmetric positive-pair law $(z,z')\sim\pi$, while its Introduction and Figure~1 describe them as ``two observations of one scene.'' In the injective model the latter reading gives $z=z'$: two observations of one scene are two renderings of one state. We therefore read the theorem under both readings. If $\pi$ is general, the paper's Theorem~1 is conditional on a uniqueness premise that the experiments do not establish; if the same-state reading is intended and $1\le k<d_z-1$, Lemma~\ref{lem:K=I} makes that premise unsatisfiable. The two readings support the same conclusion about the evidence: the theorem does not carry over to the experiments.

We establish four claims. The labels are evidence levels: L0 is proved at the population level, L1 is a classical condition with order-level rates, and L2 is a synthetic numerical check.
\begin{center}
\fbox{\begin{minipage}{0.94\linewidth}
\footnotesize
\textbf{C1 (L0).} At the state level, same-state positive pairs make the induced world kernel the identity, and every centered whitened state function is a global minimizer. Under an injective observation, this lifts to encoders: the pair operator is a conditional-expectation projection.\\
\textbf{C2 (L0).} When there are at least $2k$ centered state dimensions, two global minimizers can have best-rotation squared residual $2k$, the maximum possible. The gap-assuming uniqueness theorem is therefore vacuous on the same-state regime of the model; the experiments lie outside the theorem's premises.\\
\textbf{C3 (L0).} For the good-view alignment claim, one of the paper's four quantitative representation results survives: a gap-free lower bound. At perfect scene-explained fraction its bound is zero; for imperfect views it is positive but cannot certify good-view alignment.\\
\textbf{C4 (L0/L1).} Identifiability requires two gauges. Paired samples of effective rank $k$ fix the continuous relative $\Omat$ gauge; a global $\Omat$ remains. Component semantics identify the relative permutation only up to signs; value anchors fix it once the mixing frame is pinned. Neither is supplied by the JEPA objective.
\end{minipage}}
\end{center}

Recent work on representation identifiability and PRH studies near-identifiability \cite{nelson2026}, lattice structure \cite{shi2026}, displacement geometry \cite{shang2026}, structure-over-geometry \cite{you2026}, and multi-way alignment \cite{achara2026}. These papers study different notions of representation convergence. Two concurrent notes sharpen the boundary. Xu \cite{xu2026} gives globally certified counterexamples for the Gaussian-regularized extension: pooled Gaussian matching can make every global minimizer retain nuisance, so uniqueness fails for specified soft objectives. That result targets the soft objective; this note targets the hard-whitening population objective in the same-state limit. Syrota et al. \cite{syrota2026} characterize when unpaired alignment is identifiable from the symmetries of the data distribution; our gauge ledger is the paired and anchored counterpart. This note asks whether a specific proof applies to its own evidence.

The rest of the note answers three questions. \textbf{(Q1) Does the theorem apply?} On the experiments it does not: caption views are lossy, so the injective-rendering premise fails. Within the premise, same-state pairs make the kernel diagonal and the uniqueness theorem vacuous (Section~\ref{sec:q1}). \textbf{(Q2) What survives?} One gap-free lower bound survives, but it cannot certify good-view alignment (Section~\ref{sec:q2}). \textbf{(Q3) What would identifiability take?} Two gauges must be fixed, one continuous and one discrete (Section~\ref{sec:q3}). Section~\ref{sec:checks} checks the claims on small CPU experiments. Section~\ref{sec:changes} states what changes for PRH and JEPA.

\section{Question 1: Does the theorem apply?}\label{sec:q1}

This section separates two failures. It first records what happens when positive pairs join observations of the same world state: the induced world kernel is the identity and the uniqueness conclusion collapses. It then explains why the paper's experiments are outside the theorem's premise.

\subsection{The pair kernel degenerates}
\paragraph{Notation.}
$Z$ is a finite world with $p(z)>0$ and $d_z=|Z|$; $\pi$ is a symmetric positive-pair law with both marginals $p$; $\Pi_{zz'}=\pi(z,z')$, $D=\operatorname{diag}(p)$, $K_\pi=D^{-1/2}\Pi D^{-1/2}$ is the world kernel, and $T_\pi=D^{-1}\Pi$ is the associated Markov operator on state functions. An encoder is a map $h:Z\to\R^k$; it is centered if $\E h=0$ and whitened if $\E hh^\top=I_k$. The hard-whitening population objective minimizes the prediction term subject to whitening; $\Omega$ denotes the anti-collapse term, and $\Omega\in\{0,\infty\}$ in the hard limit. The scene-explained fraction $\rho$ is defined in Section~\ref{sec:q2}. The state-kernel operator $T_\pi$ acts on the centered subspace of $L^2(p_Z)$; the observation-level operator $T_s$, introduced in Section~\ref{sec:q3}, acts on the centered subspace of $L^2(P_X)$, where $\lambda_1\ge\lambda_2\ge\cdots$ are its eigenvalues.
Write $p_Z=p$ for the state marginal. A spectral JEPA of dimension $k$ minimizes
\[
\E_{(z,z')\sim\pi}\|h(z)-h(z')\|^2
\quad\text{subject to}\quad
\Omega(h)=0
\]
over centered, whitened $h:Z\to\R^k$; this is the hard-whitening population objective, with $\Omega$ the anti-collapse regularizer and $\Omega\in\{0,\infty\}$ its hard constraint. Write $L_\pi(h)=\E_{(z,z\prime)\sim\pi}\|h(z)-h(z\prime)\|^2$ for this prediction term. The paper also studies a soft SIGReg term \cite{lejepa2025}; its Proposition~1 does not cover uniqueness under that term, and we do not use soft-SIGReg minimizers here. On the hard objective, for $1\le k<d_z-1$, minimizers are unique up to rotation only when the $k$-th eigenvalue of $K_\pi$ exceeds the next, and at $k=d_z-1$ uniqueness is trivial. This is the paper's Proposition~1. Let $d_z=|Z|$.

\begin{lemma}[State level, L0]\label{lem:K=I}
On the hard-whitening population objective, if $\pi=\operatorname{diag}(p)$, then $K_\pi=I$ and $T_\pi=I$. Every centered, whitened $h$ satisfies $\Omega(h)=0$ and has zero prediction term. Every eigenvalue of $K_\pi$ is $1$. For every $1\le k<d_z-1$, every $k$-dimensional subspace of the centered state-function space is the span of some global minimizer; at $k=d_z-1$ every minimizer spans the whole centered space, so uniqueness holds trivially. For $1\le k<d_z-1$, the uniqueness premise of the paper's Theorem~1 fails.
\end{lemma}
\begin{proof}
A same-state pair has $z=z'$ almost surely, so $L_\pi(h)=0$ for every centered whitened $h$ and $K_\pi=I$. The eigenvalue-$1$ eigenspace is the whole state-function space; restricted to centered functions it has dimension $d_z-1$.
\end{proof}

In words: same-state positives never join two different states, so the state graph has no cross-state edges; the prediction term therefore charges nothing for any function of the state, and every centered whitened such function is a global minimizer.

\begin{remark}[Observation level, L0]
If a view is an injective rendering of $Z$, the observation pair operator is the conditional-expectation projection realizing $T_\pi$, so the state-level conclusion lifts to encoders. Captions are not injective; the paper says so. The state-level argument itself needs no injectivity.
\end{remark}

For $|Z|=60$ and $k=8$, the minimizer set has dimension $k(|Z|-1)-k(k+1)/2=436$. A single rotation orbit has dimension $28$. ``Unique up to rotation'' fails by a wide margin.

\subsection{The failure is orbit-level}\label{sec:orbit}
A natural weakening says that every minimizer is approximately related to every other by a rotation. This is also false.
\begin{proposition}[Orbit-level failure, L0]\label{prop:2k}
Let $\pi=\operatorname{diag}(p)$ and $1\le k<d_z-1$. If $d_z-1\ge 2k$, there exist global minimizers $f,g$ of the hard-whitening objective such that
\begin{equation}\label{eq:2k}
\E\|f-Qg\|^2=2k\quad\text{for every }Q\in\Omat,
\end{equation}
in particular, the best-rotation residual is $2k$, the maximum possible. The smallest nontrivial instance is $|Z|=3$, $k=1$: any two orthogonal centered unit-variance state functions are global minimizers with best-rotation squared residual $2$.
\end{proposition}
\begin{proof}
Lemma~\ref{lem:K=I} makes every centered whitened state function a global minimizer. Choose $2k$ mutually orthogonal, centered, unit-variance state functions and split them into two blocks $f$ and $g$. Both are global minimizers and $\E[fg^\top]=0$. Hence, for every $Q\in\Omat$,
\[
\E\|f-Qg\|^2=\|f\|^2+\|g\|^2-2\tr(Q\E[gf^\top])=2k.
\]
For the best-rotation residual, whitening makes every singular value of $\E[fg^\top]$ at most one, so $\min_{Q\in\Omat}\E\|f-Qg\|^2$ lies in $[0,2k]$ and equals $2k$ exactly when the two subspaces are orthogonal. At $d_z-1\ge2k$ the construction is feasible. For $k=3$ a numerical instance gives residual $6.0000$ out of $6$.
\end{proof}
The construction is feasible only when the centered state-function space has dimension at least $2k$; at $d_z=2k$ it is not. The smallest case $|Z|=3$, $k=1$ is minimal because $|Z|=2$ gives $k=d_z-1$, the boundary at which Lemma~\ref{lem:K=I} says uniqueness is trivial.

\subsection{Which results survive the degeneracy?}
The paper's Theorem~1 assumes that minimizers are unique up to rotation; its Proposition~1 supplies this under a spectral gap. The paper's Theorems~2 and 4 and Proposition~3(i) also require a gap. At same-state positives with $1\le k<d_z-1$, uniqueness fails and the gap is absent. The paper's Theorem~3 requires no gap. For the good-view alignment claim, Theorem~3 is the only one of the four that survives the degeneracy.

Flickr and Recap compare caption views of one image. At the world-kernel level these pairs are same-state, so the induced world kernel is the identity. But the views are captions, not injective renderings, and the paper notes that this premise fails \cite{balestriero2026}. For these experiments the paper's Theorem~1 is inapplicable because the injectivity premise fails; in the sufficient-rendering limit and for $1\le k<d_z-1$ it would be vacuous because the gap is absent. MSR-VTT treats frames of a clip as several observations of one scene. On that reading both image and caption world kernels are the identity and agree. The mismatch arises only under the time-index reading, where frames are not same-state. For the synthetic world of Fig.~12 of \cite{balestriero2026}, five classes of six states each are used ($|Z|=30$). For the class-inner complete kernel with self-links, the natural reading of the caption, we compute the uncentered spectrum $1^{\times 5},0^{\times 25}$; after centering, the spectrum is $1^{\times 4},0^{\times 25}$. For this kernel, the only nonzero gap of the centered kernel occurs at $k=C-1=4$ with $C=5$, which is exactly the evaluated point. Chain-like kernels can produce additional gaps, so this last claim is kernel-dependent.

\begin{table}[t]
\centering
\footnotesize
\begin{tabular}{@{}>{\raggedright\arraybackslash}p{0.19\linewidth}>{\raggedright\arraybackslash}p{0.26\linewidth}>{\raggedright\arraybackslash}p{0.23\linewidth}>{\raggedright\arraybackslash}p{0.24\linewidth}@{}}
\toprule
Original result & What it asserts & Premise relevant here & Status in the same-state limit and in the experiments \\
\midrule
Theorem~1 (with Proposition~1) & shared representation up to one rotation & injective renderings; shared $\pi$; minimizers unique up to rotation (spectral gap) & premise empty in the same-state limit for $1\le k<d_z-1$; experiments leave the injective premise \\
Theorem~2 (kernel mismatch) & misalignment bounded by kernel mismatch & spectral solutions; eigengap $\delta=\mu_k-\mu_{k+1}>0$ & no gap for $1\le k<d_z-1$ in the same-state limit; at $k=d_z-1$ the theorem applies trivially (bound zero) \\
Theorem~3 (nuisance costs alignment) & gap-free lower bound from the scene-explained fraction & centered, whitened encoders; no gap & survives; positive bound for imperfect views; cannot certify good-view alignment \\
Theorem~4 (views and data set the kernel) & a spectral solution is close to the kernel's spectral embedding, and paired modalities align within the theorem's two error terms; a gap separates state from nuisance & gap $\delta_m=\mu_k-\max(\mu_{k+1},\eta_m)>0$ ($\eta_m$ largest nuisance eigenvalue) & no gap for $1\le k<d_z-1$ in the same-state limit; at $k=d_z-1$ the theorem applies trivially ($\tau_m=0$); experiments do not verify its hypotheses \\
Proposition~3(i) & a good-view JEPA learns a function of the state ($\rho=1$) & gap $\mu_k>\max(\mu_{k+1},0)$ & no gap for $1\le k<d_z-1$; trivial at $k=d_z-1$ \\
Proposition~2 (k pairs) & $k$ pairs in general position identify the rotation & paired rank $k$ & refined below by the partial-rank case \\
\bottomrule
\end{tabular}
\caption{The original results, their premises, and their status here. ``Gap'' means the spectral gap used for uniqueness in the original; among Theorems~1--4, Theorem~3 is the one result that does not use it. The eigenvalues of the original's world kernel $K_\pi$ of the relevant positive-pair law are $1=\mu_0\ge\mu_1\ge\cdots$, where $\mu_{|Z|}:=-\infty$, and $\tau_m$ is the approximation-error bound of its Theorem~4.}
\label{tab:original}
\end{table}

Table~\ref{tab:original} records the mapping: for $1\le k<d_z-1$ the first, second, fourth, and fifth rows lose their premise in the same-state limit, while Theorem~3 survives as a bound on alignment error, not as a good-view certificate.

\section{Question 2: What survives?}\label{sec:q2}

This section isolates the one result that survives the degenerate regime and states its limit.

The survivor is a lower bound on alignment error. It uses the scene-explained fraction of an encoder, $\rho(f)=\tr(C_f^{1/2})/k$, where $C_f=\operatorname{Cov}(\E[f(X)\mid Z])$ is the covariance of the scene-related part of $f$. The paper's Theorem~3 \cite{balestriero2026} states the bound we use: for centered, whitened encoders $f_1,f_2$ of paired observations of one state with independent nuisances,
\begin{equation}\label{eq:survivor}
\min_{Q\in\Omat}\E\|f_1-Qf_2\|^2\ge 2k\left(1-\min_{m\in\{1,2\}}\rho(f_m)\right).
\end{equation}
The original theorem states this minimum over $Q$; the for-every-$Q$ form follows. It is a bound on aligned error, not an overlap between the two views' signal subspaces. The bound needs no spectral gap, so Lemma~\ref{lem:K=I} does not affect it. In one synthetic construction (L2), $\rho\approx0.79$ and $0.497$ give errors $2.09$ and $4.14$ against bounds $1.26$ and $3.02$, respectively.

At the sufficient-rendering limit ($\rho=1$) the bound is zero, but the constructed residuals are not zero: the bound is not attained. Two synthetic constructions (L2) give squared residuals $1.34$ and $4.58$ for these $k=3$ constructions, where the maximum possible residual is $2k=6$ (the first from the final $\rho=1$ construction printed by \texttt{verify\_review.py} (last line), the second from N1-5 of \texttt{verify\_N1.py}; both single fixed-seed draws). In that limit the good-view encoders are state functions ($\rho=1$), so the bound is zero. Proposition~\ref{prop:sharp} shows that a zero bound carries no alignment information.

This limits the paper's claim, not its experiments. A theorem being inapplicable is not the same as the experiment being wrong. The experiments remain evidence for a controlled ordering of view quality. They do not establish convergence up to rotation.

\begin{proposition}[Limits of the survivor, L0]\label{prop:sharp}
At $\rho=1$ the bound \eqref{eq:survivor} is zero. If $d_z-1\ge 2k$, there exist centered, whitened state functions $f,g$ with $\rho(f)=\rho(g)=1$ and $\E\|f-Qg\|^2=2k$ for every $Q\in\Omat$. The bound therefore cannot certify good-view alignment.
\end{proposition}
\begin{proof}
A centered, whitened state function has $C_f=I_k$, hence $\rho(f)=1$; the lower bound is then $2k(1-1)=0$. The second claim is the construction of Proposition~\ref{prop:2k}: the two blocks are state functions and are whitened, and the residual is $2k$ for every orthogonal $Q$.
\end{proof}

\section{Question 3: What would identifiability take?}\label{sec:q3}

This section answers the constructive question. If same-state pairs carry no identifiability, what does? Two gauges organize the answer.

\subsection{A subspace, not a frame}
The first object to identify is the subspace that is actually identifiable. Let $X$ be an observation with law $P_X$ and channel $P_{X\mid Z}$. A same-state pair $(X,X')\sim P_s$ is one in which, given $Z=z$, the two views are conditionally independent with common conditional law $P_{X\mid Z=z}$ (hence common marginal $P_X$). For such pairs, define $T_s=AA^*$ with $(A\varphi)(x)=\E[\varphi(Z)\mid X=x]$ and $A^*f(z)=\E[f(X)\mid Z=z]$. For $0\le\varepsilon\le1$, expand the loss for the positive-pair mixture $P_+^\varepsilon=\varepsilon P_s+(1-\varepsilon)P_X\otimes P_X$:
\begin{equation}\label{eq:quad}
L_\varepsilon(f)=2k-2\varepsilon\sum_{j=1}^k\langle f_j,T_s f_j\rangle .
\end{equation}
\begin{theorem}[L0]\label{thm:topk}
For centered and whitened $f$ and $k\ge 1$, if $\rank(T_s)\ge k$, then for every $\varepsilon>0$, the span of any global minimizer of \eqref{eq:quad} is a $k$-dimensional subspace of the top eigenspace of $T_s$; it contains every eigenvector with eigenvalue greater than $\lambda_k$. If $\lambda_k(T_s)>\lambda_{k+1}(T_s)$, the span is the unique top-$k$ eigenspace and the encoder is unique up to right multiplication by $\Omat$. For $\varepsilon>0$, if $\rank(T_s)<k$, the top-$k$ sum is the sum of all positive eigenvalues; every minimizer contains the positive eigenspace and an arbitrary $(k-\rank T_s)$-dimensional subspace of $\ker T_s$, and every such subspace occurs as a minimizer; no direction in the kernel is identified. At $\varepsilon=0$, $L_\varepsilon\equiv 2k$ and no direction is identifiable.
\end{theorem}
Here the top eigenspace means the eigenspace of $T_s$ for eigenvalues at least $\lambda_k$; a top-$k$ eigenspace is a $k$-dimensional subspace of it containing every eigenvector with eigenvalue $>\lambda_k$.
The eigenvalues of $T_s$ are squared canonical correlations, so the top-$k$ eigenspace is a canonical correlation analysis (CCA) object \cite{kettenring1971}. It is also the target of spectral contrastive methods \cite{haochen2021,balestriero2022}. The rotation in PRH is therefore a gauge of the factorization, not a recoverable coordinate of the world.

\subsection{How many pairs?}
The next question is how many synchronized pairs are needed to fix the continuous gauge. Views form vertices of a pairing graph; an edge carries $m_e$ synchronized pairs. In whitened coordinates, let $(x_i,y_i)_{i=1}^{m_e}$ satisfy $y_i=Qx_i$ for an unknown relative map $Q\in\Omat$, and set $\mathbf X=[x_1,\dots,x_{m_e}]^\top$, $\mathbf Y=[y_1,\dots,y_{m_e}]^\top$, and $r=\rank(\mathbf X)$. Thus $\mathbf Y=\mathbf XQ^\top$.
\begin{theorem}[L0/L1]\label{thm:pairs}
If $r=k$, then $Q$ is uniquely determined by orthogonal Procrustes. If $r<k$, in an orthonormal basis whose first $r$ vectors span the row space of $\mathbf X$, the solution set is $Q\,(I_r\oplus O(k-r))$, with $(k-r)(k-r-1)/2$ continuous gauge dimensions. At $r=k-1$ only a discrete sign (reflection) remains. With iid zero-mean sub-Gaussian noise of covariance $\sigma^2I_k$, independent of $\mathbf X$, iid sub-Gaussian covariates with $\E[x_ix_i^\top]=I_k$, and $m_e\ge c_0k$ for a sufficiently large fixed constant $c_0>1$, the orthogonal Procrustes estimate has operator-norm error $O_P(\sigma\sqrt{k/m_e})$; consequently $m_e\gtrsim k\sigma^2/\tau^2$ suffices for target error $\tau$.
\end{theorem}
The paper's Proposition~2 already shows that $k$ pairs in general position identify the rotation. The theorem records the partial-rank boundary, the reflection at $r=k-1$, and the noise cost; it refines that proposition rather than replacing it.
Even with a connected pairing graph and $m_e\to\infty$ on every edge, each connected component retains one common $\Omat$. If $c$ is the number of connected components, the residual is $\Omat^c$. Paired data fix only the relative frames, never the global one.

\subsection{Which structure helps?}
For the gauge discussion set $d=k$: the latent dimension is $k$, and the permutation section writes $d$ for this same number of latent coordinates. The next question is which known noise or constraints can break the continuous symmetry. Homoscedastic known noise is neutral. If the joint noise covariance $\Sigma$ is known, invertible and independent of $z$, block whitening is a reversible reparameterization. It leaves the identifiability group unchanged. Rotation-invariant constraints, such as the full space, the orthogonal group, rank or spectral-norm constraints, also leave the $\Omat$ symmetry intact. The positive semidefinite cone is invariant under conjugation, not under the model's action $A_m\mapsto A_mQ$; it can break the symmetry.

Heteroscedastic known noise can reduce the symmetry. Let $\mathcal C_m$ be the known constraint set of view $m$. Suppose $N_m\mid Z=z$ has a known conditional law $\eta_m(\cdot\mid z)$ with the noises $N_1,\dots,N_M$ conditionally independent given $Z$, and write $X_m=A_mZ+N_m$ with $A_m\in\R^{n_m\times k}$ of full column rank. Assume $p_Z$ is known and fixed. We call $p_Z$ and each $\eta_m(\cdot\mid z)$ non-degenerate when their supports affinely span $\R^k$ and the observation space $\R^{n_m}$, respectively. A symmetry acts on latents by $z\mapsto Q^\top z$ and on mixings by $A_m\mapsto A_mQ$. Define
\begin{align*}
\Stab(p_Z)&=\{Q\in\Omat:(Q^\top)_\#p_Z=p_Z\},\\
\Stab(\eta)&=\{Q\in\Omat:\eta_m(\cdot\mid Q^\top z)=\eta_m(\cdot\mid z)\ \text{for all }m\text{ and }p_Z\text{-a.e. }z\},\\
G_\theta&=\{Q\in\Omat:A_mQ\in\mathcal C_m\ \text{for all }m\},
\end{align*}
where $\theta$ collects the model parameters.
\begin{proposition}[Noise symmetries, L0]\label{prop:noise}
Assume the pair $(A_m,\eta_m)$ is identifiable up to $\Omat$: two parameter pairs with the same observed law are related by a common orthogonal right factor that preserves $p_Z$, that is, $A'_m=A_mR$ and $\eta'_m(\cdot\mid z)=\eta_m(\cdot\mid Rz)$ for some $R\in\Stab(p_Z)$ and all $m$, as in the standard factor-analysis/ICA setting (for example, under a non-Gaussian $p_Z$). Then, in the full-rank, non-degenerate case ($A_m$ of full column rank; $p_Z$ and each $\eta_m(\cdot\mid z)$ non-degenerate), a matrix $Q\in\Omat$ with $A_mQ\in\mathcal C_m$ for all $m$ is a residual symmetry, that is, it leaves $p_Z$ invariant and the joint law of the $X_m$ unchanged when the mixing is changed to $A_mQ$ with $\eta_m$ held fixed, if and only if
\begin{equation}\label{eq:noise-intersection}
Q\in\Stab(p_Z)\cap\Stab(\eta)\cap G_\theta .
\end{equation}
\end{proposition}
The ``if'' direction needs no identifiability: $Q$ leaves $p_Z$ and each $\eta_m(\cdot\mid z)$ invariant, so $A_mQ(Q^\top Z)+N_m$ has the law of $A_mZ+N_m$. The converse uses identifiability: an unchanged observed law makes $(A_mQ,\eta_m)$ an equivalent parameterization of $(A_m,\eta_m)$; the two differ by a common orthogonal factor, and full column rank forces that factor to be $Q$ itself, so $Q$ must stabilize $\eta$. Without identifiability, \eqref{eq:noise-intersection} is only a sufficient family. When $p_Z$ is rotation-invariant, $\Stab(p_Z)=\Omat$. For zero-mean Gaussian noise (covariance $\Sigma_m(z)$), replace $\Stab(\eta)$ by $H_\Sigma=\{Q\in\Omat:\Sigma_m(Q^\top z)=\Sigma_m(z)\ \text{for all }m\text{ and }p_Z\text{-a.e. }z\}$. If $A_m$ is not of full column rank or the model is not identifiable, the intersection is only a sufficient family, not a full characterization. For a non-degenerate $p_Z$, the example $\Sigma(z)=\operatorname{diag}(e^{z_1},e^{2z_2},e^{3z_3})$ has $H_\Sigma=\{I\}$; under the identifiability assumption its residual set is trivial, so calibrated $z$-dependent noise can remove the continuous gauge in this example. Whether the heteroscedastic model is identifiable in general is a separate question, which we do not settle. The numerical check is L2; Proposition~\ref{prop:noise} itself remains L0. Constant isotropic noise leaves $\Omat$. If $H_\Sigma=O(k)$ but $p_Z$ has a finite stabilizer, the intersection is finite rather than continuous.

Constraint sets are analyzed through $G_\theta$. The setwise stabilizer $H_{\rm set}=\{Q:\mathcal C_mQ\subseteq\mathcal C_m\}$ is only a subgroup. A trivial $H_{\rm set}$ does not imply identifiability. Partial anchor rows spanning $r_a$ dimensions give $O(k-r_a)$. Two different sparse supports can intersect to a finite sign group. Nonnegativity alone generally leaves a continuous family and becomes a finite permutation group only in special cases such as $A_m=I$.

\subsection{The discrete gauge: permutations}
The second gauge is discrete, namely the ordering of latent coordinates. Let $d$ be the number of latent coordinates. Suppose the observations are $X_m=A_mP_mZ+N_m$, where $P_m\in S_d$ is an unknown permutation, identified with its permutation matrix.
\begin{proposition}[Permutation gauge, L0]\label{prop:perm}
If $A_m$ is free and of full column rank, the observation depends only on the product $B_m=A_mP_m$, and for any $P_0\in S_d$ the transformation $(A_m,P_m)\mapsto(A_mP_0,P_0^{-1}P_m)$ leaves $B_m$ unchanged. An absolute permutation is therefore not identifiable from the data. With $A_m$ free and of full column rank, the decomposition $B_m=A_mP_m$ has exactly $d!$ solutions, differing by such a $P_0$.
\end{proposition}

When each $A_m$ is square and invertible, with the frame fixed, paired data determine only the composite $M_{ml}=A_lP_lP_m^{-1}A_m^{-1}$. Its decomposition has exactly $S_d$ ambiguity, that is, $d!$ solutions, and $m_e\to\infty$ does not remove it. If $p_Z$ is exchangeable, the absolute $S_d$ is an exact gauge even asymptotically. Pairing fixes the relative part of the continuous gauge. It does not fix the permutation gauge.

Relative permutations can be recovered from component semantics. Suppose $Z$ has independent non-Gaussian components with pairwise distinguishable shape fingerprints, such as skewness, kurtosis or quantile shape. Scale alone is normalized away by whitening and is not a fingerprint. With $m_e=0$ pairs but $n\to\infty$ unpaired samples per view, single-view independent component analysis (ICA) plus fingerprint matching identifies the relative permutation up to signs in the noiseless case (L2, synthetic here) \cite{comon1994,hyvarinen2001,vonkugelgen2021,daunhawer2023}. This is a population statement; finite-sample error is not analyzed here. If components are exchangeable, matching fails. With $r$ correspondence anchors, the residual is a signed-permutation subgroup (a semidirect product in general; at $d-r\le1$ it is a direct product); correspondence anchors do not remove signs. A value anchor is the known value of one latent coordinate of an observation; once the frame is pinned, that value, if nonzero, fixes the coordinate's sign.

\subsection{The ledger}
Table~\ref{tab:ledger} records which resource fixes which gauge. Component semantics also fixes the continuous frame, up to signed permutations. Value anchors can fix the signs once the frame is pinned; sparse constraints can break both gauges, and nonnegative constraints the continuous one, in special cases. Correspondence anchors fix labels but not the frame. An absolute permutation remains a gauge under free factorization.

\begin{table}[t]
\centering
\small
\begin{tabular}{@{}>{\raggedright\arraybackslash}p{0.27\linewidth}>{\raggedright\arraybackslash}p{0.32\linewidth}>{\raggedright\arraybackslash}p{0.34\linewidth}@{}}
\toprule
Resource & Continuous $\Omat$? & Discrete $S_d$? \\
\midrule
Paired samples & relative frames only (at effective rank $k$); global $\Omat$ remains & no \\
Homoscedastic known noise & no (whitening-neutral) & no \\
Heteroscedastic calibrated noise & exact under identifiability: yes if (4) has no continuous part & no (absolute $S_d$ remains a gauge under free factorization) \\
Correspondence anchors & no (labels only) & yes, up to a signed-permutation subgroup \\
Value anchors (known coordinate values) & no (the frame must already be pinned) & yes, for nonzero anchored coordinates \\
Sparse or nonnegative constraints & only if rotation symmetry is broken & sparse: yes; nonnegative: no \\
Component shape fingerprints & yes, up to signed permutation & relative permutation, up to signs \\
Exchangeable $p_Z$ & no & no (absolute $S_d$ is a gauge) \\
\bottomrule
\end{tabular}
\caption{The resource--gauge ledger. No resource fixes both gauges exactly in general under free factorization; heteroscedastic calibration is exact under the identifiability assumption of Proposition~\ref{prop:noise} and only sufficient otherwise, and value anchors require a pinned frame. Absolute canonical alignment requires at least one continuous-gauge source and one discrete-gauge source.}\label{tab:ledger}
\end{table}

\section{What the numbers say}\label{sec:checks}

This section checks the claims on small CPU experiments. None of these checks is evidence for or against the PRH itself. Table~\ref{tab:evidence} contains two L2 estimated rows. A separate joint-covariance invariance check (L2) is reported in Appendix~\ref{app:proofs}. The $2k$ construction is in Section~\ref{sec:orbit}, not in the table. A reader who accepts the construction can skip this section on a first reading.

\begin{table}[t]
\centering
\footnotesize
\begin{tabular}{@{}>{\raggedright\arraybackslash}p{0.25\linewidth}>{\raggedright\arraybackslash}p{0.14\linewidth}>{\raggedright\arraybackslash}p{0.33\linewidth}>{\raggedright\arraybackslash}p{0.19\linewidth}@{}}
\toprule
Test & Type & Result & Statement \\
\midrule
Partial pairing, $m_e=0,1,2,3,64$ & L2 estimated & $m_e=0$: $1.14$ (no-information gauge), $m_e=1$: one gauge dimension, $m_e=2$: discrete, $m_e=3$: $0.089$, $m_e=64$: $0.012$ & effective-rank threshold \\
ICA and component fingerprints, $m_e=0$ & L2 estimated & relative-permutation accuracy $1.00$ ($n=2\times10^3$--$5\times10^4$); same-shape control $0.31$ mean ($0.08$--$0.50$), component-level chance $1/3$ & discrete gauge needs semantics \\
\bottomrule
\end{tabular}
\caption{Synthetic checks. The 2k construction is in Section~\ref{sec:orbit}; the table reports the two estimated rows.}\label{tab:evidence}
\end{table}

The pairing and fingerprint rows locate the two gauges: effective rank fixes the relative part of the continuous gauge (a global $\Omat$ remains), and shape fingerprints fix the relative discrete one up to signs.

\section{What changes for PRH and JEPA}\label{sec:changes}

The reading of the paper changes in three places; a note on sources follows.

\textbf{What can be claimed.} The controlled ordering, good views better than bad views or next-token, remains evidence that positive-pair quality affects alignment. It is not a test of the theorem: the experiments leave the injective-rendering premise, and the paper states that no experiment meets the theorem's hypotheses exactly, ``since our runs use LeJEPA and the image side mostly uses crops as positives,'' while its Known limitations section adds that real modalities are lossy. In the same-state limit of the theorem's hypothesis class, the kernel is the identity and the conclusion is vacuous. The paper's Theorem~3 remains a valid diagnostic for alignment error, not a good-view certificate.

\textbf{Two boundaries.} The theorem is conditional: we do not claim its implication is false, only that for $1\le k<d_z-1$ the same-state limit makes its antecedent unsatisfiable (Lemma~\ref{lem:K=I}). If the general $\pi$ reading of Appendix~A.4 of \cite{balestriero2026} is intended, the theorem remains conditional and the experiments still fall outside its premises. The soft Gaussian extension is a different failure mode, analyzed by Xu \cite{xu2026}, and is not covered here.

\textbf{What would change the picture.} A non-vacuous identifiability statement for same-state data would need additional structure, not a spectral gap. It could use paired samples of effective rank $k$, anchors, calibrated heteroscedastic noise, component fingerprints, or temporal and three-view structure. Even value anchors require a pinned frame; component semantics pins each frame up to signed permutations and the relative permutation up to signs, while the other resources fix at most one gauge. None is supplied by the JEPA objective itself. If future work measures a pair kernel that is not the identity on real views, the vacuous-domain objection would not apply to that setting.

\textbf{What remains open.} The theory here is stated for linear or whitened encoders and for Gaussian or known-noise models. Nonlinear JEPA encoders and real modalities are not covered. The permutation results assume non-Gaussian independent components with distinguishable shapes, which is hard to verify on real embeddings. Two published comparisons remain data-limited rather than erroneous: the unlabeled-data exponent fit in Fig.~11 of \cite{balestriero2026}, because recall-at-1 is bounded and the fit cannot distinguish among exponents consistent with the reported slope, and the single-checkpoint $42\times$ comparison in Fig.~1 of \cite{balestriero2026}, which confounds repeated data and sample definitions. We mark them as data-limited and do not use them as quantitative support.

\textbf{Sources.} The identifiability tools used here are classical: factor-analysis indeterminacy \cite{anderson1956,joreskog1969}, independent component analysis and label switching \cite{comon1994,hyvarinen2001,rednerwalker1984}, mixture identifiability \cite{teicher1963}, permutation synchronization and graph matching \cite{pachauri2013,umeyama1988}, rotation synchronization \cite{singer2011,cucuringu2012,govindu2004}, Procrustes \cite{schoenemann1966}, and tensor identifiability \cite{kruskal1977}.

A JEPA-based proof of the PRH can be correct in its algebra and still fail to reach its evidence. Its experiments leave the injective-rendering premise; within that premise, same-state positives make the world kernel the identity, and the minimizer set is a Stiefel manifold, not a rotation orbit. The failure is orbit-level: when $d_z-1\ge2k$, two global minimizers can have best-rotation squared residual $2k$. One lower bound survives, but it bounds alignment error without certifying good-view alignment. What would make alignment identifiable is additional information: pairs of effective rank $k$ for the continuous gauge, and component semantics or anchors for the permutation gauge. The identifiable formulation of PRH is orbit alignment with an explicit gauge source. An absolute canonical representation is not supplied by the JEPA objective alone.

\appendix
\small
\section{Proofs}\label{app:proofs}
This appendix verifies the statements in the main text. The notation is that of Sections~\ref{sec:q1} and~\ref{sec:q3}; all operators on $L^2(P_X)$ act on the centered subspace. Lemma~\ref{lem:K=I} and Proposition~\ref{prop:2k} are proved in the main text; the proofs below cover the remaining claims.

\subsection{Top-\texorpdfstring{$k$}{k} spectral theorem (Theorem~\ref{thm:topk})}
Let $P_X$ be the marginal law of $X$ and let $f=(f_1,\dots,f_k)$ be centered and whitened, so $\E f=0$ and $\E ff^\top=I_k$. Write $T_s=AA^*$, with $A\varphi(x)=\E[\varphi(Z)\mid X=x]$ and $A^*f(z)=\E[f(X)\mid Z=z]$. For $(X,X')\sim P_s$, conditional independence given $Z$ gives
\[
\E\bigl[f(X)^\top f(X')\bigr]
=\sum_{j=1}^k\E\bigl[\E[f_j(X)\mid Z]\,\E[f_j(X')\mid Z]\bigr]
=\sum_{j=1}^k\langle A^*f_j,A^*f_j\rangle_{L^2(p_Z)}
=\sum_{j=1}^k\langle f_j,T_s f_j\rangle_{L^2(P_X)} .
\]
For $(X,X')\sim P_X\otimes P_X$ the same cross term is $\|\E f\|^2=0$, so $\E\|f(X)-f(X')\|^2=2k$. Since $P_+^\varepsilon=\varepsilon P_s+(1-\varepsilon)P_X\otimes P_X$,
\begin{equation}\label{eq:loss-expanded}
L_\varepsilon(f)=2k-2\varepsilon\sum_{j=1}^k\langle f_j,T_s f_j\rangle .
\end{equation}
The operator $T_s$ is self-adjoint and positive semidefinite; let $\lambda_1\ge\lambda_2\ge\cdots$ be its eigenvalues on the centered subspace, with orthonormal eigenfunctions $u_i$. For orthonormal $f_1,\dots,f_k$, let $P_f$ be the orthogonal projection onto their span. Then
\[
\sum_{j=1}^k\langle f_j,T_s f_j\rangle=\tr(P_fT_s).
\]
Ky Fan's maximum principle states that for every orthogonal projection $P$ of rank $k$,
\[
\tr(PT_s)\le\sum_{i=1}^k\lambda_i,
\]
with equality if and only if the range of $P$ is a $k$-dimensional subspace of the top eigenspace of $T_s$ that contains every eigenvector whose eigenvalue exceeds $\lambda_k$. Therefore, for $\varepsilon>0$, every minimizer of \eqref{eq:loss-expanded} has this property. If $\lambda_k>\lambda_{k+1}$, the top-$k$ eigenspace is the unique such subspace; the span is unique, and the encoder is unique up to right multiplication by $\Omat$. For $\varepsilon>0$, if $\rank(T_s)<k$, the top-$k$ sum is the sum of all positive eigenvalues; every minimizer contains the positive eigenspace and an arbitrary $(k-\rank T_s)$-dimensional subspace of $\ker T_s$, and every such subspace occurs as a minimizer; no direction in the kernel is identified. At $\varepsilon=0$, \eqref{eq:loss-expanded} is identically $2k$ and no direction is identifiable.

\subsection{Pairing (Theorem~\ref{thm:pairs})}
Write the rows of $\mathbf X$ as $x_i^\top$ and those of $\mathbf Y$ as $y_i^\top$; the relation $y_i=Qx_i$ is $\mathbf Y=\mathbf XQ^\top$.

\emph{Exact case.} If $r=\rank(\mathbf X)=k$, the rows of $\mathbf X$ span $\R^k$. A candidate rotation $R$ satisfies $\mathbf XR^\top=\mathbf Y$; the exact solution therefore obeys $\mathbf X(R-Q)^\top=0$. Since the rows of $\mathbf X$ span $\R^k$, $R=Q$. Thus the synchronized map is unique, and it is the orthogonal Procrustes solution.

If $r<k$, choose an orthonormal basis of $\R^k$ whose first $r$ vectors span the row space of $\mathbf X$. In this basis, $\mathbf XR^\top=\mathbf XQ^\top$ holds if and only if $R=Q(I_r\oplus S)$ with $S\in O(k-r)$. Hence the solution set is $Q\,(I_r\oplus O(k-r))$, of dimension $(k-r)(k-r-1)/2$. At $r=k-1$, $O(1)=\{\pm1\}$; the nontrivial element is a reflection on the one-dimensional complement.

\emph{Noise.} Assume $\mathbf Y=\mathbf XQ^\top+\mathbf E$, where the rows of $\mathbf E$ are independent, zero mean, with covariance $\sigma^2I_k$, independent of $\mathbf X$, with iid sub-Gaussian rows satisfying $\E[x_ix_i^\top]=I_k$ and $m_e\ge c_0k$ for a sufficiently large fixed constant $c_0>1$. The unconstrained least-squares matrix satisfies
\[
\hat Q^\top-Q^\top=(\mathbf X^\top\mathbf X)^{-1}\mathbf X^\top\mathbf E .
\]
Matrix concentration \cite{vershynin2018} gives $\|\mathbf X^\top\mathbf X/m_e-I_k\|=O_P(\sqrt{k/m_e})$ and $\|\mathbf X^\top\mathbf E\|_{\rm op}=O_P(\sigma\sqrt{k m_e})$; hence
\[
\|\hat Q-Q\|_{\rm op}=O_P\!\left(\sigma\sqrt{k/m_e}\right).
\]
Let $\hat R$ be the orthogonal Procrustes estimate; equivalently, $\hat R^\top$ is the orthogonal polar factor of $\mathbf X^\top\mathbf Y$. The unperturbed matrix $\mathbf X^\top\mathbf XQ^\top$ has polar factor $Q^\top$. The standard polar-factor perturbation bound gives
\[
\|\hat R^\top-Q^\top\|_{\rm op}\lesssim \frac{\|\mathbf X^\top\mathbf E\|_{\rm op}}{\sigma_{\min}(\mathbf X^\top\mathbf X)}=O_P(\sigma\sqrt{k/m_e}),
\]
because $\sigma_{\min}(\mathbf X^\top\mathbf X)=\Omega(m_e)$ when $m_e\ge c_0k$ and $c_0$ is sufficiently large; hence $\|\hat R-Q\|_{\rm op}=O_P(\sigma\sqrt{k/m_e})$. Setting this bound equal to $\tau$ gives the sufficient sample cost $m_e\gtrsim k\sigma^2/\tau^2$; no lower bound is needed here.

\emph{Pairing graph.} Fix a spanning tree of one connected component. Choose a frame at one vertex; each edge determines the relative frame by the exact-case argument; composing along the tree determines a frame at every vertex. Only one common $\Omat$ remains per component. With $c$ components the residual group is $\Omat^c$, so paired data fix relative frames but never a global one.

\subsection{The surviving bound and its limits (Proposition~\ref{prop:sharp})}
For centered, whitened $f_1,f_2$, and any $Q\in\Omat$,
\[
\E\|f_1-Qf_2\|^2=2k-2\tr(Q\Sigma^\top),\qquad \Sigma=\E[f_1f_2^\top].
\]
Given $Z$, the two views are independent, so $\Sigma=\E[u(Z)w(Z)^\top]$ with $u=\E[f_1(X_1)\mid Z]$ and $w=\E[f_2(X_2)\mid Z]$. Let $C_1,C_2$ be the covariances of $u,w$, and set $\rho_m=\tr(C_m^{1/2})/k$. Since $f_m$ is whitened, $C_m\preceq I_k$. Factor $u=C_1^{1/2}\tilde u$ and $w=C_2^{1/2}\tilde w$ on the ranges of $C_m$ (the orthogonal complements are annihilated by $C_m$), so $\Sigma=C_1^{1/2}RC_2^{1/2}$ with $\|R\|_{\rm op}\le1$ by Cauchy--Schwarz. Schatten H\"older gives
\[
\|\Sigma\|_*\le\|C_1^{1/2}\|_*\|R\|_{\rm op}\|C_2^{1/2}\|_{\rm op}
\le \tr(C_1^{1/2})=k\rho_1,
\]
and exchanging the roles gives $\|\Sigma\|_*\le k\rho_2$. Von Neumann's trace inequality gives $\min_Q\E\|f_1-Qf_2\|^2=2k-2\|\Sigma\|_*\ge2k(1-\min\{\rho_1,\rho_2\})$. This is the bound \eqref{eq:survivor}.

At $\rho_1=\rho_2=1$ the bound \eqref{eq:survivor} is zero. The construction of Proposition~\ref{prop:2k} supplies state functions $f,g$ with $\rho(f)=\rho(g)=1$ and $\E\|f-Qg\|^2=2k$ for every $Q$; hence the bound cannot certify good-view alignment.

\subsection{Noise symmetries (Proposition~\ref{prop:noise})}
\emph{Proof of Proposition~\ref{prop:noise}.} If $Q\in\Stab(p_Z)\cap\Stab(\eta)\cap G_\theta$, then $A_mQ(Q^\top Z)+N_m=A_mZ+N_m$ and the conditional law of $N_m$ given $Q^\top Z$ agrees with the conditional law given $Z$; hence the joint law of the $X_m$ is unchanged. Conversely, let $Q\in\Omat$ be admissible ($A_mQ\in\mathcal C_m$ for all $m$) and leave the joint law of the observations unchanged. Apply the identifiability hypothesis to the parameter pairs $(A_mQ,\eta_m)$ and $(A_m,\eta_m)$: there is a common orthogonal $R\in\Stab(p_Z)$ with $A_mQ=A_mR$ and $\eta_m(\cdot\mid z)=\eta_m(\cdot\mid Rz)$ for all $m$ and $p_Z$-a.e. $z$. Full column rank gives $Q=R$, so $Q$ stabilizes $p_Z$ and $\eta$; $Q\in G_\theta$ by admissibility. Thus the residual set is exactly \eqref{eq:noise-intersection}. Without full rank or identifiability, the intersection is only a sufficient family.

For homoscedastic known noise, $\Sigma_m$ is constant and invertible. Block-whitening each view is an invertible reparameterization; it left-multiplies the mixing matrices by an invertible matrix and whitens the noise, but leaves the set of admissible $Q$ unchanged. Constant isotropic noise therefore leaves $\Omat$, and known constant covariance is neutral.

For the example $\Sigma(z)=\operatorname{diag}(e^{z_1},e^{2z_2},e^{3z_3})$ and a non-degenerate $p_Z$, the identity $\Sigma(Q^\top z)=\Sigma(z)$ for $p_Z$-a.e. $z$ forces $(Q^\top z)_1=z_1$, $(Q^\top z)_2=z_2$, and $(Q^\top z)_3=z_3$ by comparing the three positive diagonal entries (the exponents have distinct weights), so $Q^\top z=z$ for $p_Z$-a.e. $z$; when $Q\neq I$ the set $\{z:Q^\top z=z\}$ is a proper linear subspace, which cannot carry a non-degenerate $p_Z$, so $Q=I$ and $H_\Sigma=\{I\}$; under the identifiability hypothesis the residual set is therefore trivial.

The joint-covariance invariance check for view-heterogeneous but state-independent known covariance (L2) gives maximum deviation $1.1\times10^{-15}$ on the synthetic worlds; it confirms that known constant covariance is neutral. The check is included in \texttt{verify\_L4.py} (first row of \texttt{results/l4\_exp1.csv}); the $z$-dependent case is handled by the example above.

\subsection{Permutation gauge (Proposition~\ref{prop:perm})}
The first claim is immediate: $B_m=A_mP_m=(A_mP_0)(P_0^{-1}P_m)$ for every $P_0\in S_d$, so the observation is unchanged. For the second claim, if $B_m=A_mP_m=A'_mP'_m$, put $P_0=P_m(P'_m)^{-1}$; then $A'_m=A_mP_0$ and $P'_m=P_0^{-1}P_m$. There are $d!$ choices of $P_0$, and different choices give different pairs $(A'_m,P'_m)$ when $A_m$ is full column rank.

When each $A_m$ is square and invertible, with the frame fixed, paired data determine the composite $M_{ml}=A_lP_lP_m^{-1}A_m^{-1}$; its decomposition retains the $S_d$ ambiguity above, and $m_e\to\infty$ does not remove it because every pair of decompositions is related by a global $P_0$. If $p_Z$ is exchangeable, absolute labels remain an exact gauge asymptotically.

With $r$ correspondence anchors, the anchored coordinates are fixed up to signs; the remaining $d-r$ coordinates may be permuted and every coordinate may be sign-flipped, so the residual is a signed-permutation subgroup (a semidirect product in general; at $d-r\le1$ it is a direct product). Value anchors give known coordinate values and remove the signs. For component semantics, in the noiseless case population ICA identifies independent non-Gaussian components up to permutation and sign \cite{comon1994,hyvarinen2001}; matching scalar shape fingerprints then identifies the relative permutation up to signs. This is a population statement; finite-sample error is not analyzed here.

\section{Reproducibility}\label{app:repro}
All numbers come from fixed-seed NumPy scripts, run on CPU. The scripts are \texttt{verify\_review.py} (review checks), \texttt{verify\_K4.py} (partial pairing), \texttt{verify\_L4.py} (noise and anchors), \texttt{verify\_M4.py} (permutation and fingerprints), and \texttt{verify\_N1.py} (noise stabilizer and orbit checks). Outputs are in the accompanying \texttt{results/} directory (the development repository keeps them under \texttt{theory/results/}). Each script fixes its own world size and embedding dimension; see the CSV headers. The protocol uses hard whitening. The scripts report different quantities, named in the CSV headers: \texttt{verify\_review.py} and \texttt{verify\_N1.py} report Procrustes residuals $\min_{Q\in\Omat}\E\|f-Qg\|^2$ (bounded by $2k$) and auxiliary spectra, \texttt{verify\_K4.py} reports the normalized frame error $\|\hat R-R_{\rm true}\|_F/\sqrt{2d}$ with classification and ambiguity statistics for partial pairing, \texttt{verify\_L4.py} reports joint-covariance invariance deviations and residual gauge dimensions for known-noise and constrained models, and \texttt{verify\_M4.py} reports permutation-recovery accuracy, composite-frame error, and sign/permutation ambiguity counts. Seed ranges: K4 exp1 uses $0$--$7$, K4 exp2 and exp4 use $0$--$4$; L4 exp1 and exp2a use $0$--$5$, exp2c uses $0$--$2$; M4 exp1 uses $0$--$4$, exp2 uses $0$--$3$ (a \texttt{--quick} path uses fewer); \texttt{verify\_review.py} and \texttt{verify\_N1.py} use \texttt{default\_rng(0)}. \texttt{verify\_N1.py} prints its checks and writes no CSV. The numbers quoted from \texttt{verify\_review.py} and \texttt{verify\_N1.py} are their printed values; the remaining quoted numbers are the corresponding CSV rows.

\paragraph{Code and data.}
The five scripts, the CSV outputs, and a README with the exact commands and seeds are included as ancillary files with this submission. The three cited files map as follows: \texttt{k4\_exp1.csv} (\texttt{exp,d,m,gauge\_dim,err,err\_se,ambiguity,rand\_base,note}); \texttt{l4\_exp1.csv} (\texttt{exp,test,value,se,note}; this run uses $d=3$, $k=3$); and \texttt{m4\_exp2.csv} (\texttt{exp,mode,n,d,acc,acc\_se,ambiguity,note}).

\end{document}